# Neural network task specialization via domain constraining

*R.O. Malashin, D.A. Ilyukhin*
*Pavlov Institute of Physiology RAS,*
*Saint-Petersburg, Makarova emb. 6.*

*Abstract*

This paper introduces a concept of neural network specialization via task-specific domain constraining, aimed at enhancing network performance on data subspace in which the network operates. The study presents experiments on training specialists for image classification and object detection tasks. The results demonstrate that specialization can enhance a generalist's accuracy even without additional data or changing training regimes – solely by constraining class label space in which the network performs. Theoretical and experimental analyses indicate that effective specialization requires modifying traditional fine-tuning methods and constraining data space to semantically coherent subsets. The specialist extraction phase before tuning the network is proposed for maximal performance gains. We also provide analysis of the evolution of the feature space during specialization. This study paves way to future research for developing more advanced dynamically configurable image analysis systems, where computations depend on the specific input. Additionally, the proposed methods can help improve system performance in scenarios where certain data domains should be excluded from consideration of the generalist network.

<u>*Keywords*</u>: neural network specialization, task-specific domain constraining, dynamic configurability.

## Introduction

Specialists [1] are neural networks trained on only a subset of the full training dataset. In image classification, a generalist model can recognize the entire range of classes in a dataset, whereas each specialist is responsible for recognizing only specific subsets of classes (e.g., only different types of mushrooms).

Though intuitive, need of specialization is not obvious as utilization of extra data for training boosts learning of general-purpose features that can enhance recognition of the classes of interest. For example, although the class "airplane" may seem irrelevant for an animal classifier, features learned from airplane images (such as shape and background) can improve the recognition of birds, especially in sky scenes where visual similarity is high. This motivates the current trend toward ever-larger models and datasets [2-4]. However, under real-world computational constraints, such scaling becomes impractical. Even without considering computational complexity, recognizing an arbitrary image in a context that is a priori unknown will be less effective than using a specialist that incorporates this context by ignoring aspects of the image that fall outside the scope of interest. The specialization approach emphasized in this work acts as a second training phase. It involves starting from a generalist model trained on the full label space, followed by fine-tuning on a constrained dataset obtained by excluding irrelevant classes (e.g., 'airplanes') from the training set. The goal of specialization is to retain useful shared features while discarding unnecessary ones, allowing the network to reallocate its capacity toward more relevant representations for the task at hand. The proposed methods can help improve system performance in scenarios where certain data domains should be excluded from consideration of the generalist network.

If we take a view in the perspective of dynamically configurable systems [5] – where different types of computations are required depending on the input data – then analyzing the training dynamics of individual specialists is important for advancing general purpose image analysis systems as well. A desirable property of dynamically configured systems is modularity: different network components are responsible for different branches of analysis and can interact with each other via minor information flow. Specialists provide one possible means of achieving such modularity, though semantic specialization is not the only approach [6]. In the simplest form of a dynamically configurable system, a generalist model can select a specialist to refine its prediction (Figure 1c).

Figure 1 provides an illustration of the main types of dynamically configurable systems, categorized by the size of the modules used within the system. The specialist-based approach can be considered intermediate between a mixture of experts (MoE) [6] (Figure 1b) and an agent-based system (Figure 1d). More details on this classification are given in the next section.

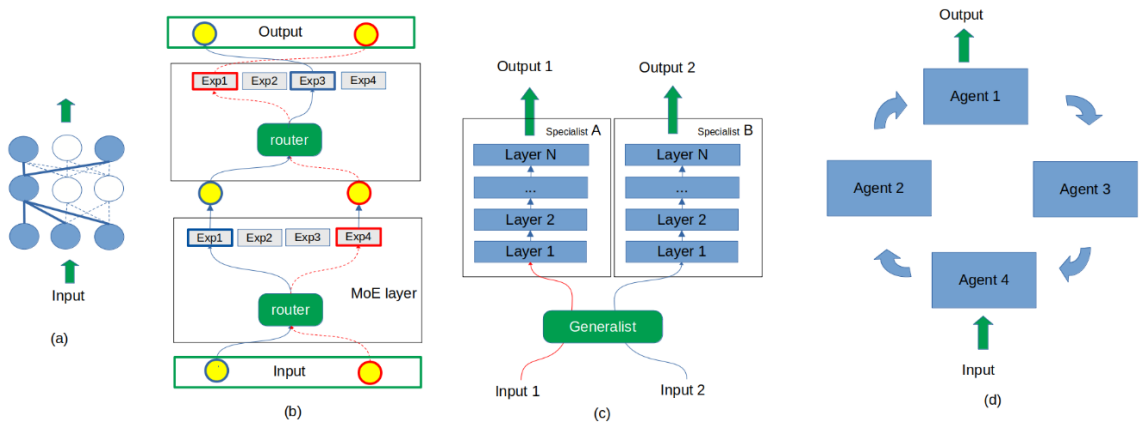

*Fig. 1. Neural network architectures with dynamically configured computations. (a) – Conditional masking of individual connections or activations. (b) – Mixture of Experts (MoE). A router determines which expert's weights will be used in the next step. (c) – Dynamic ensemble of specialists. The generalist provides an approximate solution and passes an input to the most suitable specialist. (d) – Agent-based system. A community of agents exchanges messages arbitrarily to solve the task*

Methodologically, in this paper specialization is similar to fine-tuning in transfer learning [7]. However, unlike transfer learning (or other knowledge transfer methods [8]), no new training data is introduced. Instead, performance improvement comes solely from constraining the data space the generalist was trained on. This approach also differs from dataset distillation [9], where the goal is to create a smaller training subset while preserving the model's performance.

The objective of this study is to analyze the training properties of specialized neural networks for image analysis under domain constraining. We focus on two core image analysis tasks: classification and detection. However, we believe that the insights gained from our approach may generalize to data analysis in other domains as well.

The key contributions of this work are as follows:
- We define the concept of specialization through task-specific domain constraining during training and show that specialization improves classifiers accuracy without requiring extra data or changing training regime.
- We show that specialization is only effective under semantically meaningful data clustering, and also introduce modification of classical tuning algorithm with initial phase of specialist extraction.
- We show how specialization affects internal features by applying dimensionality reduction techniques and measuring representation similarity across layers.
- We conduct experiments on detector specialization, demonstrating its effectiveness on specific data subspaces. We show that gradual domain constraining can be advantageous, and that in large detectors, freezing specific layers can lead to improved specialist performance, indicating partial modularity in the underlying architecture at the layer level.

The rest of the paper is structured as follows. Section 1 reviews related works that motivate our study. Section 2 introduces our methodology for classifier specialization via task-specific domain constraining, addressing efficient forgetting and optimization. Section 3 provides experimental justification of effectiveness of proposed methods. Section 3 extends the approach to object detection, exploring specialization effectiveness and layer-wise modularity. Finally, we discuss limitations and future research directions aimed at further refining specialization strategies.

## *1 Related works*

**Specialization.** The term specialist was introduced by Hinton et al. in their work on knowledge distillation [1]. Unlike our work they aimed building specialists for static ensembles and preventing overfitting, therefore their work doesn't give details on how individual specialist learn and do not show that specialist can improve their performance in constrained domains. Increasing neural network accuracy as a whole through specialization in practice can be challenging because training in a large label space inherently provides a regularization effect, helping to prevent overfitting [10]. In cell segmentation tasks [11] specialists are referred as models that are trained to recognize specific types of cells, while generalist generalize across different cell types.

One recent trend in computer vision is to create very big generalist model that is able to cope with multiple domains [3] and even different computer vision problems [2, 4]. We believe that achieving such an ambitious goal can be done only through explicit or implicit modularization of the model with dynamically configurable computations inside. Specialization can be thought as one way of achieving this goal.

**Specialization as the natural consequence of the training dynamics.** Learning process in deep nonlinear networks remains an open question, but studies have shown that neural network training dynamics exhibit a simplicity [12] and hypernym biases [13-14], meaning that the network initially learns to recognize broader, more general categories (like "animals") before refining its understanding of more specific categories ("husky", "spitz"). Thus, from the very beginning of training, a neural network forms a unified functional block capable of solving a coarse-grained task. The specialization of internal blocks in a neural life-learning agent [15] can emerge more or less naturally if the agent is able to select data from the same coarse category for analysis. From the perspective of single large model, different specialist agents are analogue of different modules. Thus, specialization can be interpreted as the refinement of the input data or internal representation that are routed to a concrete specialist by higher-level nodes (routers) in a dynamically configurable system. Dynamically configurable systems are beneficial when different types of computations need to be performed for different input data; these systems naturally align with specialization of their internal modules.

**Dynamically configurable systems and neural network modularity.** Neural networks implement dynamic configuration in various ways, even at the level of individual neurons (Figure 1a), without explicit modularity. For example, in [16], a method is explored for creating an "intelligent" dropout mechanism, where neurons are not deactivated randomly but based on the input data.

One well-known method for building dynamically configurable modular systems is the Mixture of Experts (MoE) [6, 17] (Figure 2b). The authors of [1] highlight that MoE has advantages over specialist ensembles, as latter require pre-defined clusters before training, whereas the MoE router learns to direct data to different experts during training, enabling more advanced specialization strategies. In large-scale language and vision-language models, MoE is often implemented at the level of individual layers or blocks [18], rather than through fully isolated specialists, resulting in different learning dynamics. Research into more effective expert mixtures is ongoing, particularly in terms of semantic coherence among experts. For instance, load balancing techniques can hinder specialization, and [19] explores strategies to mitigate this issue.

Dynamic configuration can also be implemented at higher levels of modularity. For example, [20] describes a sequential application of networks with different architectures, and [21, 22] consider classification as the sequential process of agent-environment interaction according to dynamic algorithm configuration approach [2] and the principle of least action [23, 24].

Specialization and modularity can be achieved by multiple relatively independent agents (agent society). These can be constructed, for instance, using large vision-language models that exchange information at the token level [25] (Figure 1g).

*2 Methodology*

**Definitions and problem statement.** In this section we provide analysis of classifier specialization, which we validate in the next section. We adapt the findings to detection task in Section 4.

A key aspect of specialization is defining the data space in which the model will operate. In this study, we restrict the data domain based on label space (object classes).

Let $D = \{(x_k, y_k)\}$ be the trainset, where $x_k$ is image and $y_k$ is its label. The generalist model $f_\theta$ is trained on the full dataset $D$, using the complete label set $C$ (C={0…999} for ImageNet). The objective of generalist's training is to maximize performance on the validation set $D^{Val}$.

Specialist $f_{\theta s}$ is trained with the same procedure as generalist but with the use of $D_S = \{(x_i, y_i) \in D, y \in S\}$, where $S$ is subset of all labels: $S \subset C$. And the process ($\theta \to \theta_S$) of obtaining specialized weights $\theta_S$ from generalist weights $\theta$ we call specialization. The objective of specialization is to maximize performance on the related part of validation set $D_S^{Val} = \{(x_k, y_k) \in D^{Val}, y \in S\}$, while ignoring possible performance decay on the rest of the data $D / D_S$.

We call the classes $S$ that were used for specialization to be relevant and all other classes $C / S$ to be irrelevant for the specialist $f_{\theta s}$.

**Method.** The training of a specialist is designed to achieve two key objectives: a) forget irrelevant classes that fall outside the designated subset, and b) optimize feature representations for a constrained dataset. First stage helps minimizing training objective by stopping producing probabilities of irrelevant classes.

Classifiers are usually trained with the use of cross entropy loss function:

$$L_{CE} = -\sum_j^C q_j \log p_j,$$

where **q** is one-hot encoding of the label and $p_j$ is probability of class $j$ predicted with the use of SoftMax activation function:

$$p_j = \frac{\exp(z_j)}{\sum_{k=1}^C \exp(z_k)},$$

where **z** is vector of logits.

Further we also assume that the last layer is linear that is true for most of conventional image classifiers:

$$\mathbf{z} = \mathbf{Wh} + \mathbf{b},$$

where $\mathbf{W} = [w_{ij}]$ is $|C| \times K$ weight matrix, bias vector **b** has size $1 \times |C|$, **h** is $K \times 1$ feature vector of the last layer.

To achieve efficient forgetting we propose the "extraction" procedure before fine-tuning. Extraction is achieved by composing new $|S| \times K$ matrix $\mathbf{W_S}$ and $1 \times |S|$ vector $\mathbf{b_S}$:

$$\mathbf{W_S} = \mathbf{W}[S,:],$$
$$\mathbf{b_S} = \mathbf{b}[S],$$

where $\mathbf{X}[A, :]$ means selecting rows of the matrix $\mathbf{X}$ with indexes specified by the set A. After extraction forgetting occurs automatically, while classical one-step fine-tuning approach, where the top layer is replaced with a randomly initialized new layer before fine-tuning, is less effective.

This happens because weights of the top layer are not preserved, and during their training, the useful features learned by the lower layers are disrupted and cannot be restored when training on a limited data. Notably, an established among practitioners' method, with the introduction of an intermediate training phase where only the top layer is trained while freezing all other layers, performs even worse. First phase of this procedure doesn't achieve efficient forgetting. It is true even if we freeze all the useful weights and train for many epochs. We provide experimental validation of the conclusion, and the gradients analysis during specialization of the last layer in Appendix A.

Alternatively, to extraction forgetting can be achieved by "output suppression": explicitly assigning large negative values to the logits of classes that do not belong to the specialist. However, this approach requires careful control of additional training parameters, such as the label smoothing coefficient.

We employed specialist extraction for classifiers (Section 3) and used output suppression (Section 4) for detectors.

### *3 Experiments with classifiers*

**Dataset.** We conducted experiments with the use of ImageNet [26], which contains 1000 classes. Generalist is trained with the whole dataset, while specialist sees only part of the same data.

**Model.** For our experiments it is necessary that generalist model trained on the whole dataset does not reach zero training loss, which is satisfied for parameter efficient models. We have chosen MobileNet V3 Small 1.0 [27] with 2.5M parameters to conduct extensive experiments.

**Training parameters.** Generalist models (and specialists trained from random initialization) were trained for 600 epochs using the RMSprop optimizer [28] with the following settings: batch size: 512, weight decay $10^{-5}$, dropout, RandAugment, including random pixel removal, step learning rate with noise, warm-up phase with a learning rate of $10^{-6}$ for 5 epochs, label smoothing coefficient 0.1.

With these parameters MobileNet V3 Small 1.0 reaches 67.5% accuracy on ImageNet, which is comparable to the state-of-the-art performance for this model. This ensures that any observed improvements result from specialization rather than better overall convergence of the model. Fine-tuning for specialization was performed using the Adam optimizer [29], with an initial learning rate of $10^{-4}$ for 200 epochs (without noise in the learning rate step). All other parameters remained unchanged.

**Experiment 1. Specialists for dynamically configured system.** In the first experiment, we demonstrate that specialization allows accuracy improvement even without using extra data, additional augmentation or distillation techniques. Then we build dynamic configuration system incorporating specialists.

We constructed five label subsets $S_k$ of equal size, each containing 200 ImageNet class indices. Labels from 0 to 1000 were chosen according to lexicographic order of the indexes that encode synsets (synonym sets).

The classification accuracy of generalist and specialists on different label subspaces is shown in Table 1.

*Tab. 1. The classification accuracy of generalist and specialists on different label subspaces of ImageNet. Irrelevant classes of generalist are ignored during testing (same result as for extracted and not tuned specialists)*

| classes | Generalist | Specialist |
|---|---|---|
| 0-199 | 78.5 | **80.7** |
| 200-399 | 77.9 | **80.5** |
| 400-599 | 74.0 | **75.4** |
| 600-799 | 72.3 | **73.8** |
| 800-999 | 75.7 | **77.1** |
| average | 75.7 | **77.5** |

From the table, the specialization is effective for all class subsets. It gives additional 1.8% of accuracy in average.

Based on tuned specialists, a simple dynamically configurable system can be constructed, as shown in Figure 1c. Initially, the generalist model is invoked to determine which specialist should be selected, based on the predicted label. The specialist's response then determines the final decision. Table 2 shows the result.

*Tab. 2. The classification accuracy of generalist for 1000 classes of ImageNet of generalist and dynamically configured system, when specialist refines generalist output*

| classes | Generalist | Generalist + Specialists (dynamically configured) |
|---|---|---|
| 0-1000 | 67.5 | **69.1** |

The observed improvement is very similar to the performance gains of individual specialists. This result supports the conclusion that specialists indeed perform better on a restricted data subset.

**Experiment 2. Impact of semantical coherence of label spaces.** The dominant portion of classification errors in a trained neural network are due to misclassifications between semantically related classes [21]. Since the specialist trains to correct errors of the generalist, it is reasonable to consider the specialists for semantically related groups of classes. This principle was partially applied in the previous experiment, as the ImageNet labels were selected according to synsets' indexes. Here we elaborate more on the question of the impact of semantic coherence for specialization. In [1], the authors perform clustering of the covariance matrix columns, allowing them to group together labels that are frequently predicted together. In this work, we use a static class partitioning based on the Word-Net graph [30], which effectively describes the error structure of trained networks and provides a strong intuitive interpretation of class relationships.

Using Word-Net, we formed three semantically coherent domains for specialists (similarly to [14]):

{1-7} (Miscellaneous, 80 classes).

{8} (Artifacts, 522 classes).

{9} (Animals, 398 classes).

We also formed three random domains containing an equivalent number of classes, but with class assignments made randomly, without leveraging Word-Net structure. More details on this grouping procedure can be found in [14]. As shown in Table 3, despite having fewer specialists than in the first experiment, the gain in accuracy for semantic specialists is more prominent.

*Tab. 3. Top-2 accuracy gains for specialists with randomly assigned classes and with classes assigned according to WordNet. (average improvement is measured with respect to number of classes)*

| Number of classes | Random specialist, % | Semantic specialist, % | Semantic specialist (random init), % |
|---|---|---|---|
| 80 | +0.5 | **+2.1** | -2.1 |
| 522 | +0.0 | **+1.5** | +0,7 |
| 398 | +0.2 | +2.8 | **+3.2** |
| 1000 | +0.1 | **+2** | +1.5 |

The results, shown in Table 3, indicate that for random specialists, the performance improvement is negligible, confirming that the observed gains stem from the hierarchical coherence of features within the specialist's domain.

Interestingly, we observed that randomly initialized semantic specialists (those that train from random initialization) can benefit when specializing on subsets of "animals" and "artifacts" while cannot reach generalist performance on smaller "Miscellaneous" label space. Unexpectedly, randomly initialized semantic specialist for animals' subset even outperforms semantic specialist initialized with generalist's weights. This suggests that there is room to improve the specialization procedure. We do not refer to the process of training randomly initialized specialist as specialization as it doesn't involve internal knowledge refinement.

**Feature analysis.** To investigate how feature manifold evolves during specialization we used UMAP [31] to embed features of generalist and specialist in 2D space. Specifically, we used penultimate layer features (before final linear layer) of MobileNet V3 small, which consists of 1024 element features. The result is depicted on Figure 2.

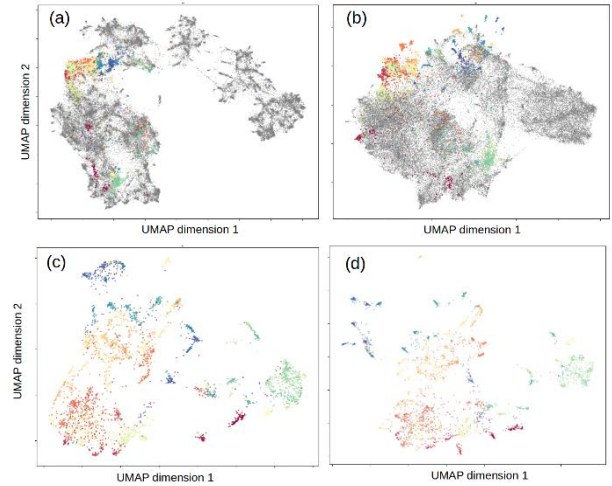

*Fig 2. UMAP embeddings of penultimate layer features (validation set). Top row – embeddings of 1000 classes (50 per class), bottom row – embeddings of 80 classes from miscellaneous superclass (50 per class). (a,c) Generalist, (b,d) – specialist (miscellaneous). Different colors of dots refer to different classes from miscellaneous superclass, gray color refers to 920 irrelevant classes that are out of the scope of the specialist*

Figure 2a shows that original feature manifold is highly structured; different "connected components" refer to different clusters of labels [14]. Specialization (Figure 2b) sacrifices that structure for irrelevant classes (grey dots), but preserves and refines structure for relevant ones (color dots). The evolution of manifold structure of relevant classes can be seen in Figure 2c and Figure 2d, where we perform UMAP embeddings for relevant classes only.

To investigate which layers are primarily affected by specialization we utilized centered kernel alignment (CKA) similarity [32], which allows more nuanced differentiation of representations than canonical correlation analysis (CCA). Figure 3 illustrates CKA similarity of generalist and specialist layers.

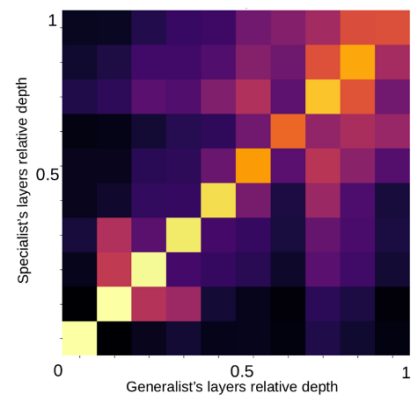

*Fig 3. CKA similarity of specialist's and generalist's layers. Specialist is trained on miscellaneous superclass*

As expected, bottom layers exhibit high similarity coefficient, which gradually decrease towards top layers.

That shows that specialization primarily affects top layers preserving low-level features.

### *Experiments with detectors*

We further study specialization in the task of object detection.

**Model.** We use the parameter-efficient YOLOv5 architecture [33], which consists of three components: a backbone (pre-trained on ImageNet for feature extraction), a neck (for feature aggregation), and a head (for predicting class labels and bounding boxes). Most experiments use YOLOv5m (21.2M parameters); for layer-freezing experiments, we use the larger YOLOv5x (86.7M parameters)

Metrics. We use the fitness metric adopted by the YOLO authors. It combines standard mean Average Precision (mAP), where mAP = TP / (TP + FN + FP), with TP defined as predictions having IoU above a threshold. mAP@0.5 uses IoU = 0.5, while mAP@0.5:0.95 averages mAP over IoU thresholds from 0.5 to 0.95. The final fitness score is computed as:

$F = 0.1 \times mAP@0.5 + 0.9 \times mAP@0.5:0.95$.

**Dataset.** We used COCO dataset [34], which consists of 80 classes grouped into 12 superclasses. Train set contains more than 118K images and 598K objects, validation set has 5K images with 37K objects. We selected "animals" superclass (63K objects for training and 2.5K for validation) for more extensive experiments. This superclass consists of 10 different animal classes: bird, cat, dog, horse, sheep, cow, elephant, bear, zebra, giraffe. We selected "horse" class to study the "second order" specialization. We employed early stopping and for this purpose we split 5000 images of COCO validation in equal parts: first part is used for validation and second part for testing, which we report in the tables. In experiments where we were interested in relative accuracies of specialists (Table 6 and Table 7) we used all 5000 images for validation and report validation accuracy.

**Training parameters.** Fine-tuning was conducted for 300 epochs, with a batch size of 16, network warm-up, and various augmentation techniques from the original project, designed for tuning on Pascal dataset [35]. Similarly, to Section 2, during specialization we excluded all images that did not contain animals or horses, as well as bounding boxes with all irrelevant classes. To train specialist from scratch we used more aggressive augmentation parameters from "scratch-high.yaml" of the yolo repository.

**Specialization.** Yolov5 outputs tensor with N×85 elements, where N is number of predicted bounding boxes. 85 elements include 4 elements that encode bounding box, 1 element of «objectness score» (which indicates the likelihood that a bounding box contains any object), and 80 values with probabilities of COCO classes. We adopted the following procedure to accomplish forgetting phase.

The YOLOv5 head does not use a SoftMax function to predict the probability distribution over all classes. Thus, we can suppress irrelevant outputs at the network's final layer, ensuring that they neither contribute to the loss function nor affect the predictions. However, the objectness score makes the forgetting phase more complex compared to classification models, because it is high for both irrelevant and relevant classes. It cannot be «switched off» for irrelevant classes other than during training.

**Experiment 1. Training specialists.** Table 4 shows the performance of generalist, randomly initialized specialist (trained from scratch) and fine-tuned specialists.

*Tab. 4. Performance of YOLOv5m on animal superclass*

| Model | Initialization | F |
|---|---|---|
| Generalist | Random | 0.662 |
| Spec. (animals) | Random | 0.677 |
| Spec. (animals) | Generalist | **0.695** |

As can be seen, specialization (tuning generalist) gives significant gain compared to generalist and training from scratch.

An animal specialist can be viewed as a generalist for a horse specialist. Thus, "horse" specialization can be performed using the animal specialist's weights as the initialization point. Table 5 shows that this gradual specialization gives significant benefit compared to direct specialization from the generalist.

*Tab. 5. Performance of YOLOv5m on horse class*

| Model | Initializa-toion | Train set | F |
|---|---|---|---|
| Generalist | Random | Whole (80 cl.) | 0,9 |
| Spec-animal | Generalist | Animals (10) | 0,913 |
| Spec-horse | Generalist | Horse (1) | 0,917 |
| Spec-horse | Spec-animal | Horse (1) | **0,936** |

This result suggests that gradually refining the input data may be more effective. This is important, as such a process could naturally align with dynamically configurable systems.

In Table 6 we report the performance improvements achieved by applying specialization for different COCO superclasses.

*Tab. 6. Performance gains from specialization of YOLOv5m. Table reports ΔF = F$^s$ - F$^G$, where F$^G$ is fitness of the generalist, and F$^s$ is fitness of a specialist*

| Domain | ΔF |
|---|---|
| Accessory (5 classes) | **+0,054** |
| Appliance (5) | +0,039 |
| Animal (10) | +0,023 |
| Electronic (7) | +0,027 |
| Food (10) | +0,003 |
| Furniture (5) | +0,045 |
| Indoor (7) | **+0,054** |
| Kitchen (7) | +0,025 |
| Outdoor (5) | +0,017 |
| Person (1) | +0,020 |
| Sports (10) | +0,047 |
| Vehicle (8) | +0,018 |

Specialization is effective among all the subsets, except the "food", where negligible improvement is achieved. Manual inspection didn't reveal anything specific about this domain, but the result aligns with [36], where the authors examined the impact of pretraining dataset similarity on fine-tuning performance: the only dataset that had no direct correlation between classification accuracy and dataset similarity (domain similarity) is Food101 [37]. A graph from [36] and sample images from the two datasets are shown in Appendix B.

**Experiment 2. Layer-wise modularity.** It can be hypothesized that certain neural network weights, trained on a larger dataset, may remain effective even when used by a specialist. To investigate this question, the following approach was taken: during specialization, some layers of the original network were frozen, meaning that the specialist retained part of the generalist's weights unchanged. As shown in Table 7, in the YOLOv5m model, freezing layers always resulted in a performance drop compared to fine-tuning without freezing. However, in the larger YOLOv5x model, freezing layers in the neck improved specialist performance.

*Tab. 7. Effect of freezing layers during specialization on animal subset. Fitness (F) is reported. The symbol "-" means that experiment wasn't conducted*

| frozen layers | YoloV5m | YoloV5x |
|---|---|---|
| - | **0,696** | 0,726 |
| Batch norm | 0,693 | - |
| 1 | 0,693 | 0,727 |
| 2 | 0,694 | 0,728 |
| 3 | 0,693 | 0,725 |
| 4 | 0,692 | 0,724 |
| 5 | 0,688 | 0,724 |
| 10-12(neck) | 0,676 | 0,729 |
| 13-20(head) | 0,693 | 0,73 |
| 13-16 | 0,676 | 0,731 |
| 13 | - | **0,733** |

### *Discussion and future work*

We believe there is enough room for further investigation of different aspects of specialization. For example, in our experiments, the label set is divided into disjoint subsets, resulting in a full decomposition of the dataset. However, fully and consistently defining relationships between classes within a single relation graph (like WordNet) is often impractical due to the multiple possible types of relationships between objects. For instance, the class "bird" belongs to the "animals", while the class "airplane" falls under "artifacts" label set. However, both can also be grouped under the broader category "flying objects", which extends beyond strict hypernymy relations. These limitations become particularly evident in cases where images of airplanes and birds depict them flying in the sky, sharing many visual features (e.g., shape, background).

How can specialists be trained to recognize different aspects of overlapping data subspaces? A key future direction is to theoretically analyze the relationship between a neural network's input, output, and the effectiveness of specialization.

Another area of research comes from possibility that the final generalist weights (with maximum accuracy) are not the best checkpoint for specialization. This is supported by the fact that in our experiments with classifiers randomly initialized "animal" specialist was able to outperform tuned extracted specialist.

### *Conclusion*

This work introduces the concept of specialization through task-specific domain constraining. We proposed a procedure that provides accuracy improvement of the image analysis networks based on constraining data distribution without requiring extra data or regularization methods. We believe the work paves a way to insights in building more efficient dynamically configured systems, including MoE and agent societies.

### *Acknowledgement*

Initial experiments were supported by the grant from the St. Petersburg Science Foundation. The experiments with detectors were supported by the State funding allocated to the Pavlov Institute of Physiology, Russian Academy of Sciences (№1021062411653-4-3.1.8).

## *Appendix A*

Here we show that training of the last layer cannot achieve effective forgetting. It is true even if we freeze all the useful weights and train for many epochs. Refer to Figure 4 to grasp additional details of this setup.

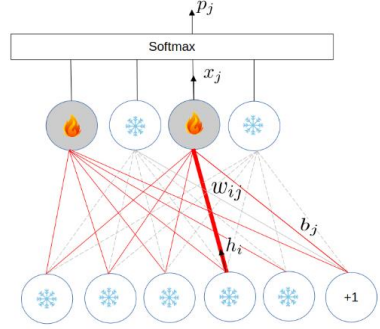

*Fig. 4. Illustration of the first stage pretraining with freezing all the important weights*

We follow the notation defined in Section 2. The derivative of the cross-entropy loss with respect to the logit has well known form:

$$\frac{\partial L_{CE}}{\partial z_j} = p_j - q_j.$$

Then let's consider the derivative of the loss with respect to the weight $w_{ij}$ (bold in Figure 4):

$$\frac{\partial L_{CE}}{\partial w_{ij}} = \frac{\partial L_{CE}}{\partial z_j} \cdot \frac{\partial (Wh + b)}{\partial w_{ij}} = (p_j - q_j)h_i = p_j h_i.$$

The last equality holds because we use only images of classes from the set $S$ to train the specialist, while only the weights corresponding to C / S classes are updated.

Thus, $w_{ij}$ will be updated according to the gradient: $\Delta w_{ij} \propto -p_i h_j$, where $p_i \geq 0$ for all $i$.

Since $h_j$ is frozen, optimization results in adopting a value of weight that minimizes:

$$w^*_{ij} = \arg\min_{w_{ij}} E_{h_j \sim D}(p_j h_i).$$

Such optimization cannot minimize all components $p_i h_j$ simultaneously, as each step will inevitably increase some $p_i h_j$ if $h_j$ take both positive and negative values for inputs from $D_S$. Using ReLU to obtain **h** ensures non-negative feature values, but, for example, MobileNet v3 employs HardSwish**,** which allows negative $h_i$. Thus, tuning the weights of the last layer cannot guarantee perfect forgetting stage that as with extracting procedure.

At the same time $\partial L_{CE}/\partial b_j = p_j$, with $p_j \geq 0$ indicating that the biases of irrelevant classes inevitably decrease though the course of training, which in theory provide the necessary forgetting, as setting a large negative value for all $b_j$, where $j \notin S$, ensures the same accuracy as extracting a specialist. However, in practical settings, the network fails to independently learn large negative biases for irrelevant classes (see Figure 5). This is due to the vanishingly small *average* probability assigned to irrelevant classes in the training set. Thus, learned weights do not generalize well on the validation set (see Figure 5a). Therefore, extraction of specialist is justified procedure to achieve effective forgetting phase. Setting large negative biases to the neurons of the irrelevant classes cannot be applied, if label smoothing coefficient is non zero (Figure 5b): label smoothing distributes a small non-zero probability mass to every class, even the irrelevant ones. When the net-work outputs near-zero probabilities for the irrelevant classes (due to the large negative biases) but the smoothed target assigns them a small positive probability, the cross-entropy loss becomes very large for these classes.

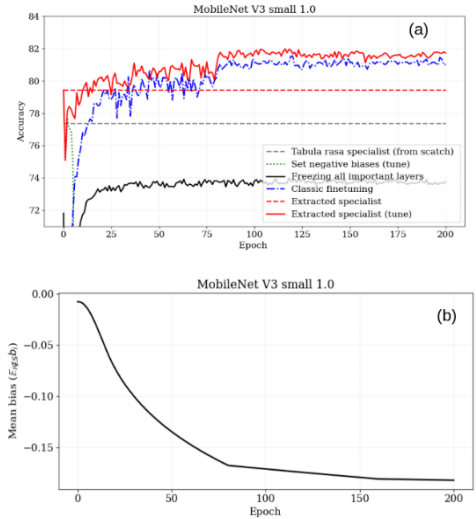

*Fig 5. Training dynamics during specialization under various settings. (a) – The most effective specialization is achieved on extracted specialist, while tuning of top layer according to Fig.2 cannot achieve accuracy of extracted specialist, showing imperfect forgetting. (b) – biases of neurons responsible for irrelevant classes decrease too slowly to compensate positive responses on every example. Specialist was trained on miscellaneous subset ImageNet (see Section 2)*

## *Appendix B*

The food specialist trained on COCO shows negligible improvement with respect to generalist, which aligns with findings of [34]. In their study, the only dataset for which no direct correlation was found between classification accuracy and dataset similarity (domain similarity) was Food101 [35]. Fig. 6 shows graph from [34] and sample images from the datasets.

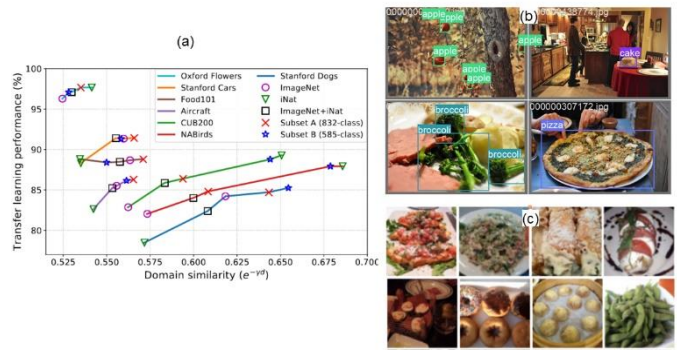

*Fig. 6. (a) - Relationship of feature transfer quality and domain similarity (between the source and target domains) [34], (b) - example images from the food category in the COCO dataset, (c) - example images from the Food101 dataset*

## *Author's information*

**Roman Olegovich Malashin** (b. 1987) graduated from Saint Petersburg State University of Aerospace Instrumentation in 2011. He received Phd degree in University of Information technology, mechanics and optics in 2014. The title of the Phd thesis was "Structural analysis of images of 3-D scenes". From 2017 he is the head of the Artificial intelligence and neural networks group in Pavlov Institute of Physiology RAS. Research interest lies in the fields of computer vision, machine learning and artificial intelligence. E-mail: *malashinroman@mail.ru*.

**Daniil Andreevich Ilyukhin** (b. 2000) graduated from Saint Petersburg State University of Aerospace Instrumentation in 2024. He currently works as a research engineer at the Pavlov Institute of Physiology. The research interest lies in the field of computer vision, machine learning and artificial intelligence. E-mail: *ne.free532@gmail.com*.